\documentclass{bmvc2k}
\usepackage[utf8x]{inputenc}
\usepackage{amsmath}[]
\usepackage{graphicx}
\usepackage{graphicx}
\usepackage{adjustbox}
\usepackage{amssymb}
\usepackage{amsmath}
\usepackage{adjustbox}
\usepackage{caption}
\usepackage{soul}
\usepackage{multicol}
\usepackage{calc}
\usepackage{floatrow}
\usepackage{listings}
\usepackage{gensymb}  
\usepackage{rotating}

\title{Deep Segmentation and Registration in X-Ray Angiography Video}

\addauthor{Athanasios Vlontzos}{athanasios.vlontzos14@imperial.ac.uk}{1}
\addauthor{Krystian Mikolajczyk}{k.mikolajczyk@imperial.ac.uk}{1}

\addinstitution{
	Dept. of Electrical and Electronic Engineering\\
	Imperial College\\
	London, UK
}
\runninghead{Vlontzos, Mikolajczyk}{Deep Segmentation and Registration }

\begin{document}
	\maketitle
	
	\begin{abstract}
		In interventional radiology, short video sequences of vein structure in motion are captured in order to help medical personnel identify vascular issues or plan intervention. Semantic segmentation can greatly improve the usefulness of these videos by indicating exact position of vessels and instruments, thus reducing the ambiguity.  We propose a real-time segmentation method for these tasks, based on  U-Net network trained in a Siamese architecture from automatically generated annotations. We make use of noisy low level binary segmentation  and optical flow to generate multi class annotations that are successively improved in a multistage segmentation approach. We significantly improve the performance of a state of the art U-Net at the processing speeds of 90fps.
	\end{abstract}
	\section{Introduction}
\begin{figure}[h]
		\begin{adjustbox}{width=\textwidth}
			
			\subfigure[ ]{\label{f1-a}\includegraphics[scale=0.2]{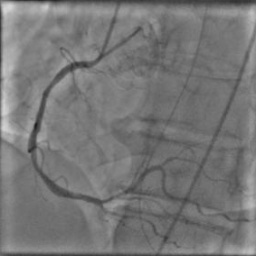}}
			\subfigure[ ]{\label{f1-b}\includegraphics[scale=0.2]{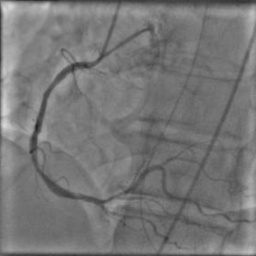}} 
			\subfigure[ ]{\label{f1-c}\includegraphics[scale=0.2]{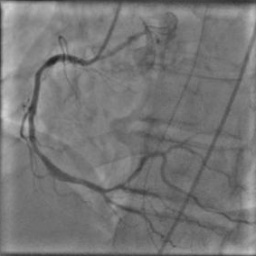}}
			\subfigure[ ]{\label{annot-d}\includegraphics[scale=0.2]{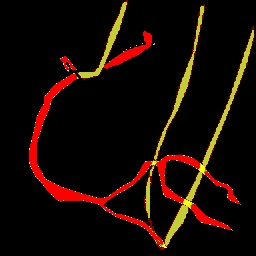}}
			\subfigure[ ]{\label{annot-e}\includegraphics[scale=0.2]{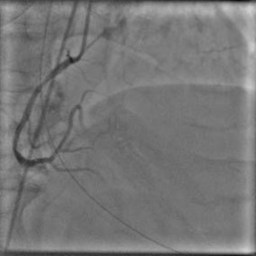}}
			\subfigure[ ]{\label{annot-f}\includegraphics[scale=0.2]{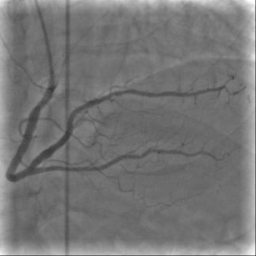}}

			\caption{Frames from a video angiogram. From left to right: (a)-(c) three consecutive frames showing significant non rigid motion due to heart beating, (d) three class segmentation into background (black), vessels (red), and catheter (yellow);  (e) frame from the same patient but a different sequence at different angle; (f) frame from a different patient.}
			\label{illustration}
			% 			\includegraphics[width=\linewidth]{annotations_copy.png}
			%\includegraphics[height=0.16\textheight]{bin_seg.png}
			% 		
			% 			\label{annotations}
		\end{adjustbox}
		
	\end{figure}

	X-ray angiography is the most used imaging modality to visualise blood vessels for interventional purposes such as stenting of stenosed vessels or for diagnostic purposes such as assessment of myocardial perfusion or stenosis grading.
	To minimise ionising radiation exposure of the patient and medical personnel during image acquisition, low power X-Rays are used resulting in noisy and low contrast images. In the context of diagnosis, the main object of interest is the vascular tree, its branchings and variations in thickness. It is therefore necessary to accurately highlight the vessels in consecutive frames to reduce the noise and improve contrast. In addition, in interventional procedures, identifying interventional instruments (catheter, wires) is also needed in order to better plan and control their positioning.  Efficiently discriminating between instruments and vessels as well as other anatomical structures that may have similar appearance is crucial during the interventions. Figure \ref{illustration}(a-c) shows an example of an angiogram sequence. Note large non-rigid motion between frames as well as the ambiguity between vessels and the catheter.  Figure \ref{illustration}(e) shows a frame from a different sequence of the same patient but taken at different scan and angle and (f) shows a different patient.  There is a significant difference in vessel as well as catheter locations in all three sequences, which we consider as independent examples.  Figure \ref{illustration}(d) shows the ground truth segmentation of the first frame.  

	%\begin{figure}[ht]
	%\begin{adjustbox}{width=\textwidth}
	%\includegraphics[width=12cm]{intro.png}}
	%\caption{Frames from a video angiogram. From left to right: Three consecutive frames showing significant non rigid motion due to heart beating, frame from the same patient but a different sequence at different angle, three class segmentation into background (black), vessels (red), and catheter (yellow); Frame from different patient {\bf can you find an example more similar to the one on the left?}.}
	%\label{illustration}
	%\end{adjustbox}
	%\end{figure}
	CNN based methods provide state of the art segmentation results but need to be trained with a large number of examples with ground truth segmentations  that are typically obtained by manual annotation by experts. The presence of noise and low contrast make this task particularly challenging and methods with handcrafted features lead to inaccurate object detection. Neural networks were demonstrated to achieve outstanding performance in vision tasks when trained from well annotated and large datasets. This is in contrast to unsupervised training where limited success has been achieved so far. We make a step towards that direction by developing an approach to automatically generate annotated examples by exploiting the knowledge of the application context and the data.

	In this paper we present a multistage architecture based on Convolutional Neural Networks for semantic segmentation of instruments and vessel tree. 
	We propose a method to automatically generate label proposals that are successively refined in a multistage process. This data is then used to train a CNN in a way that exploits spatial and temporal continuity. This results in a network capable of generating accurate segmentations. 
	%\paragraph{Contributions}
	In summary, our main contributions are: 
	\begin{itemize}
		\item We introduce a new approach for weakly supervised training of segmentation network applied to video angiograms. We propose a method that exploits low-level segmentations and optical flow to generate annotations for  multi-class video segmentation.
		%approach  for angiography X-rays simple preprocessing step to acquire automated but low accuracy binary masks of Angiography X-Rays. 
		\item We demonstrate that a convolutional  network can be trained from automatically segmented sequences with noise, and can improve the quality of these unsupervised segmentations. 
		%\item We introduce a real time CNN based multistage architecture to perform unsupervised multi-class segmentation of catheters and contrast agent in X-Ray Angiography videos.
		\item We improve the performance of the state of the art segmentation network U-Net~\cite{unet} by 20\%, processed with 90fps.
	\end{itemize}

	\section{Related work}
	\paragraph{Segmentation.} 
	Image segmentation has been a well researched topic, with new techniques arising frequently. Pixel-wise segmentation strives to create pixel masks (labels) that correspond to objects of interest in the image. One of the most common variational approaches from~\cite{Friedman2013} estimates  the background using an incremental version of EM and then subsequently subtracts it from the images. Many variational methods rely on the calculation of a Hessian matrix to acquire a vesselness index, i.e. the probability of a given pixel  belonging to a vessel. In~\cite{Hessian_segmentation},\cite{Felfelian2016} the authors construct a Hessian filter to enhance and segment coronary arteries, based on the Hessian matrix and the eigenvalues of the image. In~\cite{vessel_enh_directional_filter_bank},  a decimation free directional filter bank is used to decompose, analyse and then average the images to produce a final vesselness metric that is less sensitive to noise. In~\cite{vessel_detection_coronary} the Hessian filter is enhanced with a pipeline consisting of a guided filter, Canny edge detection and block searching in order to determine the vessels of the angiography X-Rays;~\cite{diaphragm_detection} further extends this approach to address its diaphragm misclassification. An interesting variational method comes from~\cite{layer_segmentation_vessels}, which split the images into three layers; the breathing layer, the quasi-static layer and the vessel layer based on the movement of each of the layers. The layers are then processed using morphological operators and principal component analysis to segment the vessels.    A Polygonal Path Image was introduced in \cite{Polygonal_path_image} to unify local and global curvature structures with controllable smoothness to highlight guide wires.
	
	Due to the data hungry nature of CNN approaches and the lack of annotated fluoroscopy data, the related work has been focusing on modalities like Magnetic Resonance Imaging (MRI), Computer Tomography (CT) or Positron Emission Tomography (PET). One of the most popular and effective architectures for medical image segmentation has been U-Net as introduced in~\cite{unet} or 3D-Unet ~\cite{arXiv:1606.06650}, where a symmetric convolutional network is segmenting and reconstructing  images. A residual neural network type architecture with two streams, a low resolution and a high resolution one, was investigated in~\cite{Kamnitsas16} to segment brain lesions from MRI volumes. In~\cite{Rajchl16}, an extension of the well known GrabCut~\cite{Rother04} was introduced  for pixel wise segmentation initialised by a bounding box of a of neo-natal brain lesion.
	
	\paragraph{Tracking and registration.} Tracking of blood vessels in angiography is very challenging since vessels appear and disappear as contrast agent is injected and vessel shapes change because of the heart and lung motion. Due to changes in appearance, template trackers (e.g. \cite{templ_track1}) are less robust.  Model based methods such as \cite{elastic1} regard blood vessels as active curves and deform the curves in the current frame to match the curves in the next frame. They are sensitive to noise and large change of the vessel shape. This problems led to classification-based tracking where pixels are classified in every frame as belonging to objects or background instead of transforming the appearance of objects from frame to frame. Such a tracker is robust against appearance changes and intermittent object presence. Classification-based trackers can be static e.g.~\cite{static_track} where training is performed prior to tracking  or adaptive, e.g. \cite{adaptive_track} where a classifier is built during tracking. Since our goal is to create a real-time pipeline, our preference is with static-type tracking using a convolutional neural network which can be very fast in inference mode while it can generalise well so that shape variability is accounted for.\\
	A multi-resolution registration algorithm was proposed in \cite{yang1}  where the mask image is decomposed to coarse and fine sub-image blocks iteratively and each block is rigidly registered to the live image. 
	An iterative refinement algorithm was introduced in \cite{wang} for registration in DSA. Nonrigid motion is iteratively modelled using thin-plate spline (TPS)  calculated from a set of corresponding interest points. The iterative nature of such registration algorithms leads to high computational time, which makes them difficult to use in real-time clinical applications. In \cite{2d-3d-registration} the authors use a prior CT Scan to perform a 3d to 2d registration of the coronary artery. The process however is not real time.
	
	\section{Weakly supervised segmentation}
	In this section we present our approach for generating segmented sequences that are then used to train an automatic segmentation method.  
	There is a large number of available unannotated X-ray angiography sequences. These sequences are captured after a catheter is introduced and contrast agent injected. In the first few frames, only the catheter is visible, with vessels appearing once contrast agent is injected and propagates through the vessels. The goal is to classify pixels into three labels i.e. {\em background}, {\em vessels} with contrast agent, and a {\em catheter}. 
	
	\begin{figure}[ht]
		\begin{adjustbox}{width=\textwidth}
			\includegraphics[width=\textwidth]{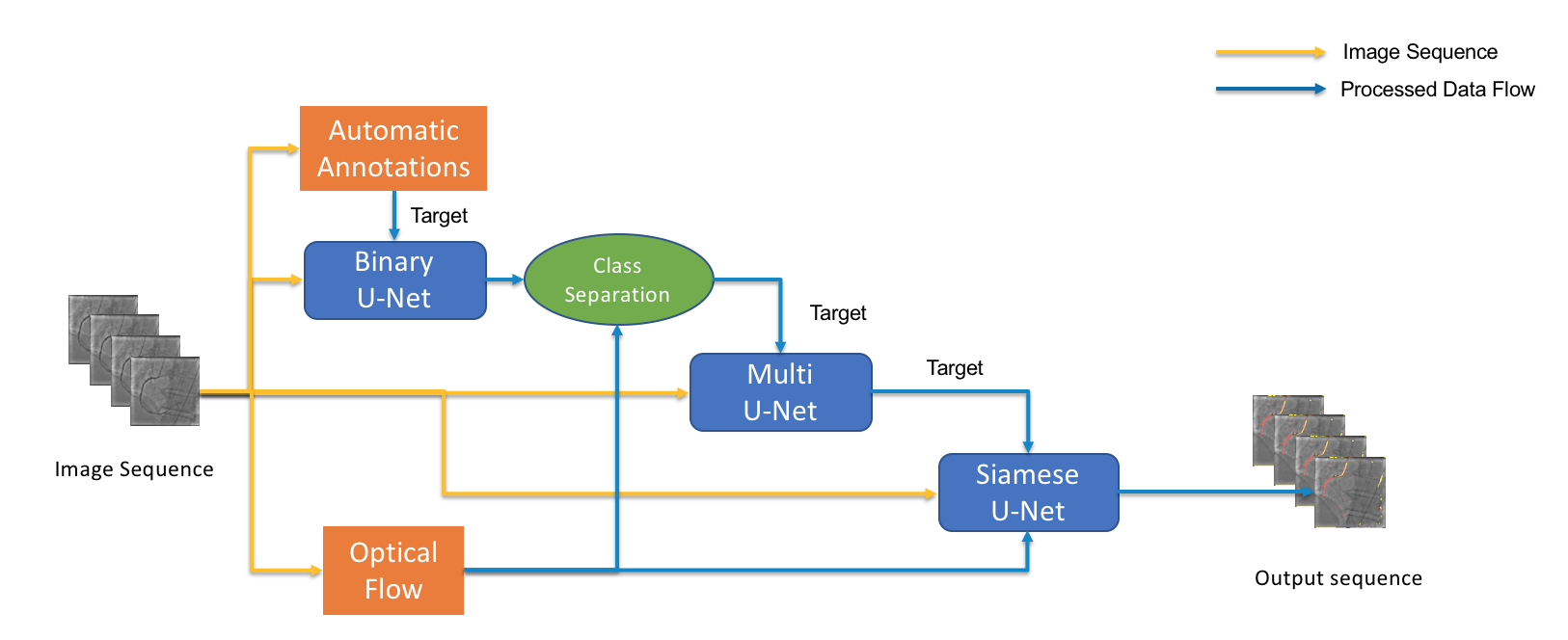}}
		\caption{Proposed training pipeline. Input: sequence of X-Ray images; Preprocessing (orange): automatic annotations and optical flow estimation; CNN based segmentation (blue), binary, multi-class and custom Siamese U-Net; Optical Flow (green) estimation for class separation. Each step of the pipeline uses the output of the previous as its target.}
		\label{pathway}
	\end{adjustbox}
\end{figure}

The proposed pipeline for annotating data and training Siamese U-Net is illustrated in Figure \ref{pathway}. We first employ a simple low-level morphological operator to generate binary masks for training a binary U-Net that discriminates between background and  other labels. Such trained U-Net outputs segmentation masks with significantly reduced noise compared to the low level processing. Motion maps, generated by coarse-to-fine  optical flow \cite{arXiv:1612.06370}, provide additional information for more accurate segmentation and discrimination between catheter and contrast agent. The binary segmentation masks are then converted into three class labels using initial frames and their motion maps. This data is then used to train a  multi class U-Net, which further  improves multi-class segmentation. Finally, a Siamese U-Net is trained using image sequences, motion maps and refined multi-class segmentation masks. The approach trains a multi-class segmentation convolutional network with an unsupervised process, otherwise weakly supervised  if a small amount of manually annotated data is used to pre-train or fine-tune the network in addition to the large volume of data that is automatically labelled.
In the following we discuss each of the components in more details. 

%The proposed method is shown  in figure \ref{pathway}. In training, it consists of three stages: The first stage generates initial estimates of the training labels  and uses them to train a U-Net as a binary classifier that produces a new candidate label identifying the pixels of interest. The resulting binary label is converted to a multiclass one using optical flow prior information. 

%In the second stage, the new multiclass labels are used to train a multiclass U-Net to produce a better estimate of the labels. Finally, in the third stage a siamese U-Net is trained with the labels produced in the previous stage incorporating motion information for better segmentation. The final siamese network can further be fine tuned to improve segmentation accuracy.

\subsection {U-Net segmentation \label{UNET}} 
Our segmentation pipeline relies on U-Net approach \cite{unet} at various processing stages, we therefore briefly explain its architecture here.
U-Net consists of two streams; a convolutional (downstream) and a de-convolutional (upstream) path. The convolutional path is based upon traditional convolutional neural network layers including spatial convolutions, max pooling and activation functions (ReLU). The outputs of consecutive layers are down-sampled with a stride of 2 while increasing the number of features at each level by 2. There are 4 levels  with 3 convolutional layers each, that generate a compressed representation of 1024 features. This is followed by a de-convolutional stream  which up-converts the feature maps while concatenating them with the corresponding ones from the convolutional stream using skip connections to preserve details. ReLU layers are used after each deconvolution. This architecture was designed to learn a segmentation process of still images and training is typically performed with binary cross-entropy loss  between original images and their segmentation masks.

%The pipeline only requires a small number of X-Ray sequences (originally developed with 1000 frames) but it also allows the use of ground truth annotations. Thus we characterise the pipeline as unsupervised with the option of semi or full supervision.

\subsection {Multi-class separation \label{PREPROC}} 
Background and motion maps are used to infer the three label segmentations masks from a video sequence.
\paragraph {Background segmentation}is performed for every  input frame. Since interventional tools and vessels with contrast agent appear darker than the surrounding pixels, we employ the black top-hat morphological operator~\cite{blackhat_reference}, defined as the difference between the closing of the image and the original one to produce approximate binary masks. The operator is used with a $9\times9$ structuring element. We eliminate small artefacts produced by the morphological operation using connected components analysis. 
The generated data still contains some noise as well as irregular vessel and catheter boundaries. Moreover there are some significant parts of vessels missing in such segmentations. We observed that a U-Net trained with such data can significantly reduce the occurrence of noise while successfully adding the missing parts. Figure~\ref{annotations}  shows an example of such segmentation.
We therefore improve the binary segmentation by using U-Net pretrained with the initial binary masks. 

% In figure \ref{PreprocImage} we exhibit two sample frames with the corresponding automatic annotations and colour coded optical flow. 

%In order to discriminate between vessels and catheters, we employ tracking of the masks generated in the first few frames of a sequence that contain  the catheter only. Tracking is based on dense optical flow computation \cite{arXiv:1612.06370} rather than non-rigid registration with features. Dense optical flow is khown to be  effective in case of neighboring frames and degrading in long range predictions. However, in case of catheters and wires it is sufficiently robust as their pixels have better contrast and are less influenced by noise. %To improve tracking, use of the segmentations generated by the CNN is made.

\paragraph {Catheter transfer. } 
The binary masks  do not distinguish between vessels and catheters.
According to standard medical practice, the angiography recording is started before introducing the contrast agent, since the latter is absorbed within a few seconds. Thus the first few frames, typically the second, as well as their binary segmentations can be used to discriminate between background and the catheter. In order to transfer the catheter label from the initial frames to those with contrast agent and visible blood vessels we use dense optical flow vectors efficiently calculated with coarse to fine approach \cite{arXiv:1612.06370}.
Motion maps are then used to transform the binary mask of frame $f_0$ with the catheter only, onto frame $f_{t}$ that contains both catheter and vascular tree. We denote the result of the non-rigid warping $f_{0t}$. The segmentation labels that appear in both, $f_t$ and  $f_{0t}$, correspond to the catheter while the binary labels in $f_t$ that find no correspondence in $f_{0t}$ are classified as contrast agent (blood vessel). Thus, we generate multi-class  labels using the binary masks and the optical flow.  Similarly to the processing of binary masks we use this data to train a multi-class U-Net and apply it to improve the segmentation results.  The multi-class U-Net is based upon the earlier binary block with the substitution of the last convolution layer with a $256\times256\times3$ layer. In addition we use categorical cross entropy (cf. Equation \ref{eq:loss}) as the loss function as opposed to the binary cross entropy from the earlier version. The input to the multi-class U-Net are the raw frames and the optical flow class separated masks as target labels.

\begin{figure}[ht]
	\begin{adjustbox}{width=\textwidth}
		\includegraphics[width=12cm]{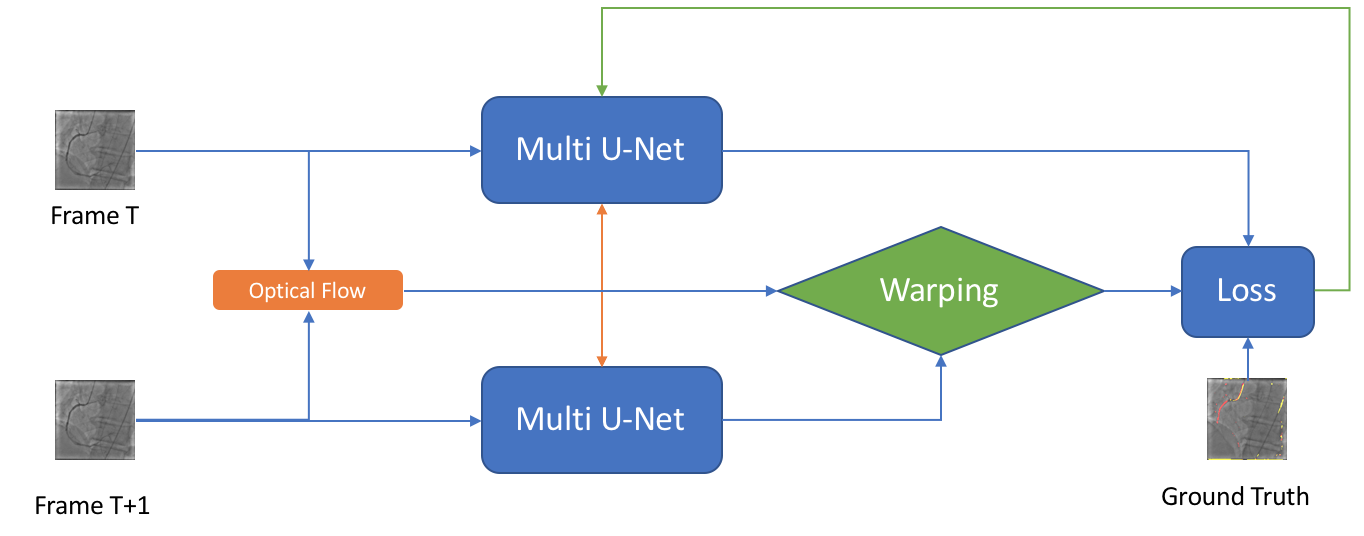}}
	\caption{Siamese U-Net; Input: pair of frames $f_t$ and $f_{t+\Delta t}$ and optical flow between the frames;  Multi-Class U-Nets in Siamese setting sharing weights.  Warping of segmentation mask of frame  $f_{t+\Delta t}$ according to Optical Flow. Categorical cross-entropy loss uses segmenation masks of both, $f_t$ and $f_{t+\Delta t}$, to update the weights. }
	\label{double_unet_path}
\end{adjustbox}
\end{figure}

\subsection {Siamese U-Net \label{SEGPATH}}
U-Net as proposed in~\cite{unet} and discussed in Section~\ref{UNET} is designed to segment static images. We  extend this approach to make use of the temporal information captured by dense motion maps. Inspired by the success of Siamese architecture in vision related tasks such as matching two images, we train a Siamese network that analyses pairs of frames from different time instances and enforces temporal consistency between their segmentations. 

%as well as frames from  temporal information as well as spatial into account during the learning process. 

The architecture is illustrated in Figure~\ref{double_unet_path} and consist of two U-Nets that share their weights. 
%We will denote each as $U_1$ and $U_2$. 
At each training iteration the Siamese U-Net takes two frames, $f_t$ and $f_{t+\Delta t}$, producing segmentations $s_{t}$ and $s_{t+\Delta t}$. The dense motion map  generated by optical flow between the two frames is used to transform  $s_{t+\Delta t}$ into $s^{\prime}_{t+\Delta t}$. In case of errorless optical flow and spatial segmentation, $s_{t}$ and $s^{\prime}_{t+\Delta t}$ should be identical. The multi-class cross entropy loss is  
\begin{equation}
\ell(f_{t},f_{t+\Delta t},y_{t})=-\sum_{c}y_{t,c} [\log(s_{t,c}) +\log(s^{\prime}_{t+\Delta t,c})]
\label{eq:loss}
\end{equation}
where $y_{t,c}$ are automatically generated segmentation labels as discussed in Section \ref{PREPROC} and index $c$ corresponds to one of the three classes in the segmentation and ground truth masks i.e. background, catheter,  and vessels.

%At this point we use the pre-calculated optical flow between $F_t$ and $F_{t+1}$, warping $U_2(F_{t+1})$ backwards towards $U_1(F_{t})$ making $U_1(F_{t})_{warped}$. Both $U_1(F_{t})$ and $U_1(F_{t})_{warped}$ are then compared to the ground truth label produced by the Multi-Class U-Net of the previous stage and both $U_1$ and $U_2$ are updated. In essence we force the networks to learn weights that would produce $F_t$ given $F_{t+1}$ and optical flow. In figure \ref{double_unet_path} we exhibit the flowchart of the proposed siamese U-Net architecture

%The aforementioned pipeline is only needed for training. For deployment only  the Double U-Net needs to be used. The inclusion of optical flow and $U_2$ lies upon the discretion of the user, however, we note that the best results are obtained through $U_1$. 

\section{Experimental results}
In this section we  evaluate our proposed approach and discuss results. We note that we have evaluated the pipeline both in its unsupervised capacity as well as the weakly supervised by injecting a small amount of manually annotated images for finetuning. 
\paragraph{Dataset}
The dataset consists of anonymised fluoroscopy X-Rays of 26 different patients.  The images were acquired during stent placement using a General Electric Innova 2000 system and stored according to standard medical protocol in DICOM format. In total the dataset includes 36000 frames corresponding to 365 distinct video sequences with an average of 98 frames each. Different sequences of the same patient were taken at different angles and stages of the procedure therefore they differ significantly as shown in Figure \ref{illustration}(c)(e)(f). Each frame is rescaled from  $512\times512$  to $256\times256$ due to memory constraints. 
\begin{figure}[ht]
\begin{adjustbox}{width=\textwidth}
	
	\subfigure[ ]{\label{annot-a}\includegraphics[scale=0.2]{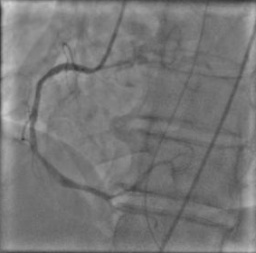}}
	\subfigure[ ]{\label{annot-b}\includegraphics[scale=0.2]{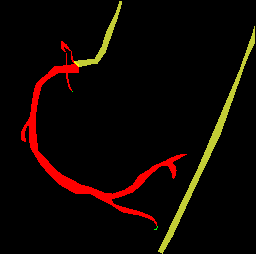}} 
	\subfigure[ ]{\label{annot-c}\includegraphics[scale=0.2]{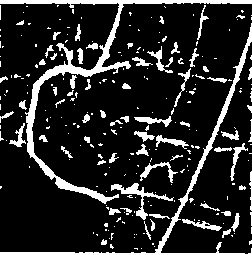}}
	\subfigure[ ]{\label{annot-d}\includegraphics[scale=0.2]{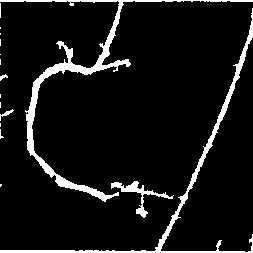}}
	\subfigure[ ]{\label{annot-e}\includegraphics[scale=0.2]{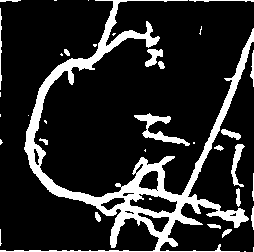}}
	\subfigure[ ]{\label{annot-f}\includegraphics[scale=0.2]{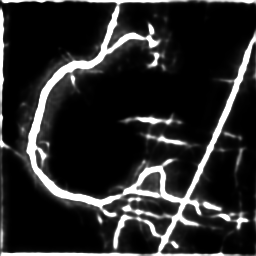}}
	\subfigure[]{\label{annot-g}\includegraphics[scale=0.2]{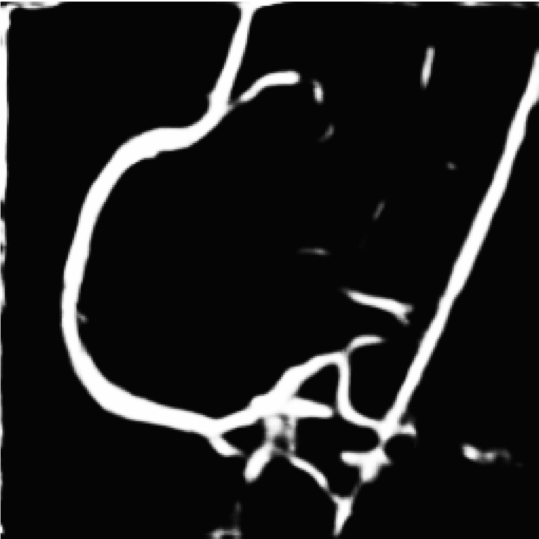}}
	\caption{ Segmentation process.  From left to right:  (a) original frame, (b)  ground truth of three colour labels, (c) Black top-hat based background segmentation,  (d) Mask filtered with connected components, (e) U-Net mask, (f) Siamese U-Net mask, (g) Fine-tuned U-Net. }
	\label{annotations}
	% 			\includegraphics[width=\linewidth]{annotations_copy.png}
	%\includegraphics[height=0.16\textheight]{bin_seg.png}
	% 		
	% 			\label{annotations}
\end{adjustbox}

\end{figure}
We randomly sampled 3000 pairs of frames not used in the test set, and used them to automatically generate annotations for training as discussed in Section \ref{QUAL} and Section \ref{QUANT}.  Furthermore, two sequences of 85 frames each, were manually annotated to serve as our test set for quantitative evaluation. Finally, we randomly selected 85 training frames for which we performed manual annotation using MIT's CSAIL LabelMe tools \cite{LabelMe}. These images were then augmented 10-fold by applying random rotations within range $[-45^\circ,45^\circ]$, and used in data augmentation experiments.

\subsection{Qualitative evaluation}\label{QUAL}
Using the approach discussed in section \ref{PREPROC} we automatically generated annotations for  3000 randomly selected frames from sequences different than those with manual ground truth. 
In order to highlight the difference between the  manual and automatic segmentation we present both in figure \ref{annotations}. The ground truth mask includes three labels i.e. background (black), vessel tree (red), and catheter (yellow). Black top-hat based segmentation mask contains significant noise, boundary irregularities and large parts of vessels as well as catheter are missing. The masks filtered with connected components seem to reduce the amount of noise but also further remove some small valid segments. This is crucial when comparing quantitative results of this simple segmentation method with and without connected component filter in Section \ref{QUANT}.

\begin{figure}[h]
\begin{adjustbox}{width=.9\textwidth}
	\subfigure[ ]{\label{Multi_Seg-a}\includegraphics[scale=0.2]{orig_fr.png}}
	\subfigure[ ]{\label{Multi_Seg-b}\includegraphics[scale=0.2]{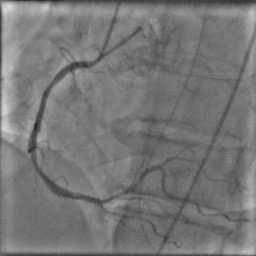}} 
	\subfigure[ ]{\label{Multi_Seg-c}\includegraphics[scale=0.2]{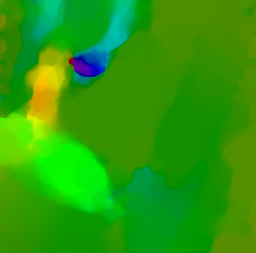}}
	\subfigure[ ]{\label{Multi_Seg-d}\includegraphics[scale=0.2]{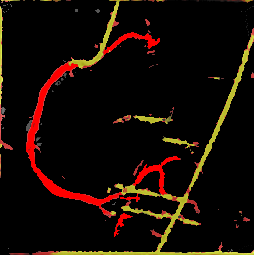}}
	\subfigure[ ]{\label{Multi_Seg-e}\includegraphics[scale=0.2]{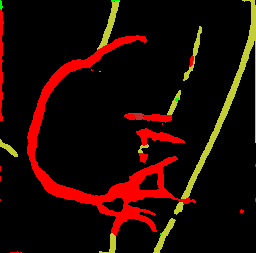}}
	\subfigure[ ]{\label{Multi_Seg-f}\includegraphics[scale=0.2]{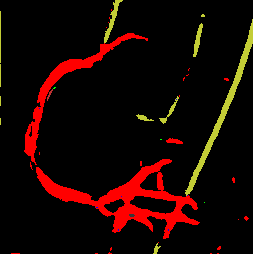}}
	\subfigure[ ]{\label{Multi_Seg-g}\includegraphics[scale=0.2]{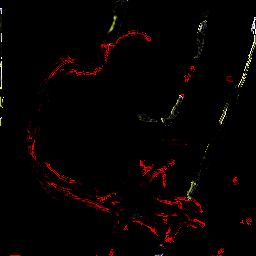}}
\end{adjustbox}
\begin{adjustbox}{width=.9\textwidth}
	\subfigure[ ]{\label{Multi_Seg-h}\includegraphics[scale=0.16]{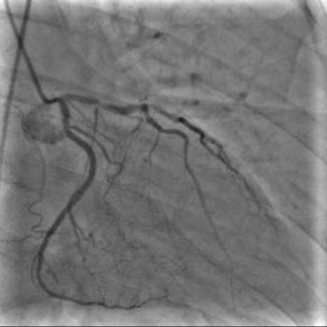}}
	\subfigure[ ]{\label{Multi_Seg-i}\includegraphics[scale=0.16]{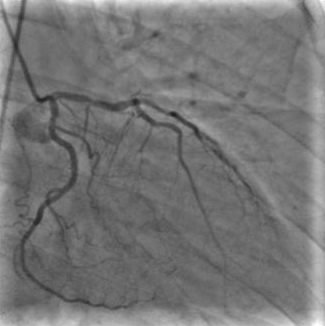}} 
	\subfigure[ ]{\label{Multi_Seg-j}\includegraphics[scale=0.16]{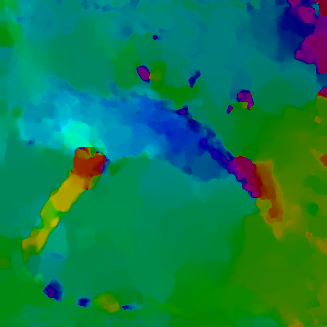}}
	\subfigure[ ]{\label{Multi_Seg-k}\includegraphics[scale=0.2]{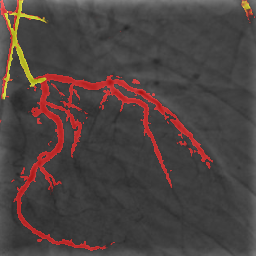}}
	\subfigure[ ]{\label{Multi_Seg-l}\includegraphics[scale=0.2]{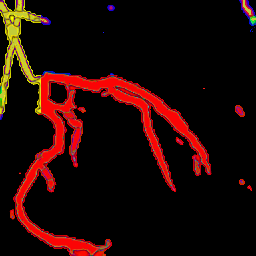}}
	\subfigure[ ]{\label{Multi_Seg-m}\includegraphics[scale=0.2]{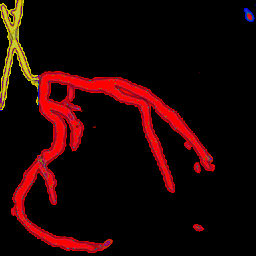}}
	\subfigure[ ]{\label{Multi_Seg-n}\includegraphics[scale=0.16]{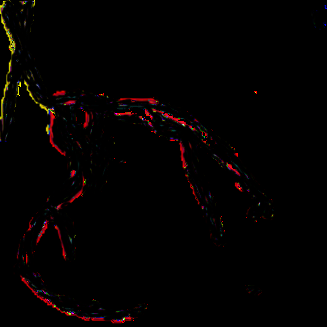}}
	\caption{
		%{\bf you can add here a frame with optical flow, small motion vectors in one frame and a frame warped with optical flow. also it would be good to have two rows with two sets of examples for different sequences.}
		Multi-class segmentation results.  (a) original frame at $t$, (b) original frame at $t+1$, (c) optical flow (HSV, hue=angle, value=magnitude), (d) optical flow based class separation,  (e) multi-class U-Net generated mask, (f) Siamese U-Net mask, (g) difference between warped mask $t+1$ and mask $t+1$. (h) to (n) results from a different sequence.}
	\label{Multi_Seg}
\end{adjustbox}
\end{figure}

%\paragraph{U-Net based segmentation} 
3000 pairs of frames with their optical flow from the automatically generated dataset were used to train the binary U-Net and the Siamese U-Net models presented in Sections \ref{UNET} and \ref{SEGPATH}. As illustrated in Figure \ref{annot-f} the Siamese U-Net is more accurate than the baseline U-Net. U-Net fine-tuned with manually annotated and augmented data provides masks with less noise and smooth object boundaries, although the same segments are successfully extracted in both, unsupervised Siamese U-Net and the fine-tuned one. 
% The rightmost image of figure \ref{binary_segmentation} has been created after fine-tuning the single U-Net with manually annotated images. As expected the semi-supervised method produces masks with less false positives but not necessarily with more true positives; in fact quantitative analysis in section \ref{QUANT} shows that the fine-tuned images are slightly less accurate.

%Following the data flow of our proposed pipeline we perform optical flow based class separation and multi class segmentation using both the simple U-Net and the Siamese U-net as described above. 

Figure \ref{Multi_Seg} compares multi-class segmentation obtained with optical flow, multi-class U-Net and Siamese U-Net. All methods were trained with automatically generated annotations without fine-tuning. Both methods provide visually comparable segmentations  therefore in the following section we perform a quantitative analysis of the segmentation accuracy.

Some of the errors appearing in the segmentation (Figure \ref{Multi_Seg-g},\ref{Multi_Seg-n}) can be attributed to the use of  simple interpolation layers instead of transposed convolutions.  Despite the use of skip connections between encoding and decoding sections of the U-Net, the use of simple up sampling layers introduces loss of details and thickening of the vessels by a few pixels. We believe that use of transposed convolutions would improve results slightly. This will be the subject of future study.

\subsection{Quantitative evaluation}\label{QUANT}
In this section we present quantitative results for our proposed segmentation approach. The comparison is done using the dice index defined between segmentation masks $X$ and $Y$
\begin{equation}
Dice=\frac{2\|X\cap Y\|}{\|X\| + \|Y\|}=\frac{2TP}{2TP+FP+FN}
\end{equation}
where $TP$ stands for the true positive number of labeled pixels, $FP$ is the false positive number and $FN$ are false negative pixels.  Dice index, however, is best suited for binary segmentation, therefore, to address this shortcoming we report per class dice index in addition to an overall score related to the binary segmentation by the proposed methods. In the multi-class case we define the binary mask as the union of the catheter and contrast agent masks. Table \ref{Quant_results_1}  shows the results for various stages of the binary segmentation pipeline noting the variant of the approach. 
%There are some interesting results that we should note that stem from the table below. To begin with the 
Interestingly, according to the dice index, the black top-hat segmentation produces more accurate masks than the connected components filter. As discussed in Figure~\ref{annot-d} the connected components filter also removes some true positive segments in the mask.  Unsupervised binary U-Net significantly improves the results. %Using data augmentation for training the U-Net gives limited improvements.
The proposed Siamese U-Net provides the best results with dice score of 0.98. Note that the multi-class labels are converted here to binary masks of background and the remaining two labels considered as one. 
\begin{table}[h]
\begin{tabular}{|c|c|}
	\hline
	\textbf{Architecture} & \textbf{Dice} \\
	\hline\hline
	Black top-hat + CC & 0.77\\
	\hline
	Black top-hat & 0.85\\
	\hline
	U-Net & 0.92\\
	\hline
	% U-Net + Augm. & 0.92\\
	%\hline
	U-Net + CC  & 0.89\\
	\hline
	Siamese U-Net & 0.95\\
	\hline
	% Siamese U-Net + Augm. & 0.90\\
	%\hline
	Siamese U-Net + CC  & {\bf 0.98}\\
	\hline
\end{tabular}
\caption{Binary segmentation results. CC stands for connected component postprocessing. Best dice score is obtained by the proposed Siamese U-Net trained in unsupervised way without data augmentation.}
\label{Quant_results_1}
\end{table}
We further evaluate our models in multi-class segmentation task and report the dice scores for each label in table  \ref{Quant_results_2}. Again, our proposed Siamese U-Net consistently improves upon the state of the art U-Net from~\cite{unet}. We test two models trained with data augmented with different transformations i.e. small random rotation \& translation in Augm\_1 vs. rotation only in Augm\_2. The latter seems to be more suited for the type of transformations that the input sequences typically include. 
Our best performing Siamese U-Net outperforms the original approach from~\cite{unet} by more than 20\%.

% \begin{table}[h]
% \begin{tabular}{|l|c|c|c|}
% \hline
% \textbf{Architecture} & \textbf{Total } & \textbf{Catheter } & \textbf{Vessels}\\
% \hline\hline
%  U-Net & 0.86 & 0.49 & 0.30\\
% \hline
%  U-Net + CC  & 0.91 & 0.44 & 0.36\\
% \hline
% Siamese  Unet + CC  & 0.91 & 0.51 & 0.50\\
% \hline
% Siamese U-Net + Augm\_1 & 0.90 & 0.61 & 0.49\\
% \hline
% Siamese U-Net + Augm\_2 & 0.90 & 0.69 & 0.54\\
% \hline
% \end{tabular}
% \caption{Multiclass segmentation results. Augm\_1 is augmentation with translations and rotations while Augm\_2 is with rotations only.}
% \label{Quant_results_2}
% \end{table}

\begin{table}[h]
\begin{tabular}{|l|c|c|}
	\hline
	\textbf{Architecture}  & \textbf{Catheter } & \textbf{Vessels}\\
	\hline\hline
	U-Net  & 0.49 & 0.30\\
	\hline
	U-Net + CC   & 0.44 & 0.36\\
	\hline
	Siamese  U-Net + CC   & 0.51 & 0.50\\
	\hline
	Siamese U-Net + Augm\_1  & 0.61 & 0.49\\
	\hline
	Siamese U-Net + Augm\_2  & {\bf 0.69} & {\bf 0.54}\\
	\hline
\end{tabular}
\caption{Dice scores for multi-class segmentation. Augm\_1 is augmentation with translations and rotations while Augm\_2 is with rotations only. Our Siamese U-Net with data augmentation significantly improves the results upon the original U-Net ~\cite{unet}.}
\label{Quant_results_2}
\end{table}

\section{Efficiency}
The proposed method was implemented in Python and the Keras framework within Tensorflow and tested with a single NVidia Titan-X GPU. In inference mode, the processing speed of U-Net trained in Siamese setup is 90 frames per second, which is suitable for real time segmentation of angiography sequences. Training time ranges from 20 min up to 8 hours when all 3000 automatically generated annotations are used for training.

\section {Conclusions}
We have introduced an approach for weakly-supervised training of a multi-class segmentation network in medical application of angiogram sequences. Our pipeline exploits automatic low level binary segmentations as well as motion vectors from optical flow to generate data annotations for multi-class segmentations thus avoiding tedious  manual segmentation of many training examples. We show that automatic annotations that include some noise can still be used to train a CNN approach and lead to improved results. We also demonstrated how to enforce the temporal consistency between segmented frames within a siamese training architecture. Multi-class U-Net trained in siamese settings   significantly improves the accuracy of segmentations compared to the original U-Net approach. Our network achieves processing speed of 90 frames per second, far beyond the speed capabilities of most of the existing interventional systems which makes it suitable for a real time applications. More experiments with different type of medical data using similar training setup will further validate this approach.
\paragraph{Acknowledgements}
The authors would like to thank Dr Tsagarakis, the chair of the scientific and ethical boards of the Evangelismos General Hospital, Athens, Greece for providing anonymized angiography sequences. 
This work was supported by EPSRC grant EP/N007743/1.%This data will not be shared by the authors without the written approval of the aforementioned hospital's scientific and ethical boards.

%\newpage
%{\bf make the references consistent, remove 'Proceedings, editions, use abbreviations eg. MICCAI'}
\bibliography{MLForIVBib}

\end{document}